\pdfoutput=1
\documentclass[11pt]{article}
\usepackage{emnlp2021}
\usepackage{times}
\usepackage{latexsym}
\usepackage[T1]{fontenc}
\usepackage[utf8]{inputenc}
\usepackage{microtype}
\usepackage{enumitem}
\usepackage{tcolorbox}
\usepackage{color,soul}
\usepackage{inputenc}
\usepackage{multirow}
\usepackage{flushend}
\usepackage{amssymb}
\usepackage{pifont}
\usepackage{todonotes}
\usepackage{url}

\newcommand{\cmark}{\ding{51}}%
\newcommand{\xmark}{\ding{55}}%
\RequirePackage{colortbl}
\definecolor{airforceblue}{rgb}{0.36, 0.54, 0.66}
\definecolor{amaranth}{rgb}{0.9, 0.17, 0.31}
\definecolor{applegreen}{rgb}{0.55, 0.71, 0.0}
\definecolor{alizarin}{rgb}{0.82, 0.1, 0.26}
\definecolor{azure}{rgb}{0.0, 0.5, 1.0}
\definecolor{cadmiumgreen}{rgb}{0.0, 0.42, 0.24}

\title{HinGE: A Dataset for Generation and Evaluation of Code-Mixed Hinglish Text}

\author{Vivek Srivastava \\
  TCS Research\\ Pune, Maharashtra, India \\
  \texttt{srivastava.vivek2@tcs.com} \\\And
  Mayank Singh \\
  IIT Gandhinagar\\ Gandhinagar, Gujarat, India \\
  \texttt{singh.mayank@iitgn.ac.in} \\}

\begin{document}


\maketitle
\setcounter{page}{1}
\begin{abstract}
Text generation is a highly active area of research in the computational linguistic community. The evaluation of the generated text is a challenging task and multiple theories and metrics have been proposed over the years. Unfortunately, text generation and evaluation are relatively understudied due to the scarcity of high-quality resources in code-mixed languages where the words and phrases from multiple languages are mixed in a single utterance of text and speech. To address this challenge, we present a corpus (\textit{HinGE}) for a widely popular code-mixed language Hinglish (code-mixing of Hindi and English languages). \textit{HinGE} has Hinglish sentences generated by humans as well as two rule-based algorithms corresponding to the parallel Hindi-English sentences. In addition, we demonstrate the inefficacy of widely-used evaluation metrics on the code-mixed data. The \textit{HinGE} dataset will facilitate the progress of natural language generation research in code-mixed languages.  
\end{abstract}

\section{Introduction}
Code-mixing is the mixing of two or more languages in a single utterance of speech or text. A commonly observed communication pattern for a multilingual speaker is to mix words and phrases from multiple languages. Code-mixing is widespread across various language pairs, such as Spanish-English, Hindi-English, and Bengali-English. Recently, we observe a boom in the availability of code-mixed data with the inflation of the social media platforms such as Twitter and Facebook. 

In past, we witness magnitude of work to address standard code-mixing natural language understanding (NLU) tasks such as language identification~\cite{shekhar2020language,singh2018language, ramanarayanan2019automatic}, POS tagging~\cite{singh2018twitter,vyas2014pos}, named entity recognition ~\cite{singh2018language}, and dependency parsing~\cite{zhang2019cross} along with sentence classification tasks like sentiment analysis~\cite{patwa2020semeval, joshi2016towards}, stance detection \cite{10.1145/3371158.3371226}, and sarcasm detection~\cite{swami2018corpus}. Unlike code-mixed NLU, natural language generation (NLG) of code-mixed text is highly understudied. Resource scarcity adds to the challenge of building efficient solutions for code-mixed NLG tasks. Evaluation of the code-mixed NLG tasks also lacks standalone resources, theories, and metrics.  

Recently, we observe a growing interest in the code-mixed text generation task. To generate the code-mixed data various techniques have been employed such as matrix frame language theory \cite{Lee2019, gupta-etal-2020-semi, dhruval_GCM_dependency_2021}, equivalent constraint theory \cite{pratapa-etal-2018-language}, pointer-generator network \cite{2018arXiv181010254I, winata-etal-2019-code, gupta-etal-2020-semi}, Generative Adversarial Networks (GANs) \cite{Gao_Bert_GAN_2019}, etc. The majority of the available datasets \cite{rijhwani-etal-2017-estimating, solorio-etal-2014-overview, patro-etal-2017-english} employed in code-mixed NLG contains noisy code-mixed text collected from social media platforms such as Twitter. These datasets also lack the sanity check for the quality of sentences, making the systems developed on these datasets vulnerable to real-world applicability.
To address the challenge of scarcity of high-quality resources for the code-mixed NLG tasks, we propose \textit{HinGE} dataset that will facilitate the community to build robust systems. The dataset contains sentences generated by humans as well as two rule-based algorithms. In Table \ref{tab:comparison}, we compare \textit{HinGE} with three other baseline datasets that can be used in the Hinglish code-mixed text generation and evaluation task.

\begin{table*}[!tbh]
\centering
\resizebox{\hsize}{!}{
\begin{tabular}{|c|c|c|c|c|}
\hline
\textbf{Dataset characteristics}                       &   \newcite{banerjee2018dataset} & \newcite{srivastava2020phinc}      & \newcite{gupta-etal-2020-semi} & \textit{\textbf{HinGE}}    \\ \hline
Parallel source sentences                     &     Only ES  &     Only ES            &       \cmark     & \cmark   \\ \hline
Human-generated code-mixed sentences        &  \cmark        &  \cmark          &      \xmark     & \cmark   \\ \hline
\begin{tabular}[c]{@{}c@{}}Multiple human-generated code-mixed \\sentences for a parallel sentence \end{tabular}        & \xmark    & \xmark      &   \xmark        & \cmark   \\ \hline
Machine-generated code-mixed sentences  & \xmark    & \xmark                    &   \cmark   & \cmark   \\ \hline
\begin{tabular}[c]{@{}c@{}}Multiple machine-generated code-mixed \\sentences for a parallel sentence \end{tabular}        & \xmark    & \xmark      &   \cmark        & \cmark   \\ \hline
\begin{tabular}[c]{@{}c@{}}Human ratings for the quality of \\generated code-mixed sentences\end{tabular} &    \xmark & \xmark & \xmark    & \cmark  \\ \hline
Dataset size   &  \begin{tabular}[c]{@{}c@{}}6733 UEU,\\6549 UHU \end{tabular} & 13,738      & 
\begin{tabular}[c]{@{}c@{}}1,561,840 PS, \\ 252,330 MGHS \end{tabular} &  \begin{tabular}[c]{@{}c@{}}1,976 PS, \\4,803 HGHS,\\ 3,952 MGHS \end{tabular}  \\ \hline
\end{tabular}}
\caption{Comparison between the various datasets available for the Hinglish NLG tasks. UEU: Unique English Utterance, UHU: Unique Hinglish Utterance, ES: English Sentences, PS: Parallel Sentences,  HGHS: Human-Generated Hinglish Sentences, MGHS: Machine-Generated Hinglish Sentences.}
\label{tab:comparison}
\end{table*}

In addition to the code-mixed NLG, the evaluation of the generated code-mixed text is a challenging task. The widely popular metrics for monolingual languages fail to capture the linguistic diversity present in the code-mixed data, such as spelling variation and complex sentence structuring. The quality ratings of the sentences generated by the rule-based algorithms in \textit{HinGE} dataset will help to develop the metrics and theories for evaluating the code-mixed NLG tasks. 
Our main contributions are:
\begin{itemize}
    \item We create high-quality human-generated code-mixed Hinglish sentences corresponding to the parallel Hindi-English sentences. Each pair of parallel sentences has at least two human-generated Hinglish sentences.
    \item In addition to the human-generated code-mixed sentences, we propose two rule-based algorithms to generate the Hinglish sentences.
    \item We demonstrate the inefficacy of five widely popular metrics for the NLG task with the code-mixed text.
    \item To develop efficient metrics for the code-mixed NLG tasks, we provide the human ratings corresponding to the code-mixed sentences generated by the rule-based algorithms.
\end{itemize}

\section{Human-Generated Hinglish Text}
\label{sec:dataset}

The scarcity of high-quality code-mixed datasets limits current research in various NLU tasks such as text generation and summarization. To address this challenge, we create a human-generated corpus of Hinglish sentences corresponding to parallel monolingual English and Hindi sentences. We use the IIT Bombay English-Hindi Parallel Corpus (hereafter \textit{`IIT-B corpus'}) \cite{kunchukuttan2018iit}. The IIT-B corpus has 1,561,840 parallel sentence pairs in English and Hindi. The English sentences are written in the Roman script, and the Hindi sentences are written in the Devanagari script. We randomly select 5,000 sentence pairs, in which the number of tokens in both the sentences is more than five to create a human-generated parallel corpus.

To create the gold-standard dataset, we employ five human annotators. Each annotator has expert-level proficiency in writing, speaking, and understanding English and Hindi languages. The objective of the annotation is to generate at least two unique Hinglish code-mixed sentences corresponding to the parallel English and Hindi sentence pairs. The annotators can also generate more than two code-mixed sentences for each sentence pair. We shuffle, pre-process, and share the sentence pairs with the annotators to generate the corresponding Hinglish sentences. A single annotator annotates each sentence pair. 

We assign 1,000 unique sentence pairs to each annotator with the following annotation guidelines: 
\begin{itemize}
    \item The Hinglish sentence should be written in Roman script.
    \item The Hinglish sentence should have words from both the languages, i.e., English and Hindi.
    \item Avoid using new words, wherever possible, that are not present in both the sentences. 
    \item If the source sentences are not the translation of each other, mark the sentence pair as ``\#''.
\end{itemize}

\begin{figure}[!tbh]
\centering
    \includegraphics[width=1.0\linewidth]{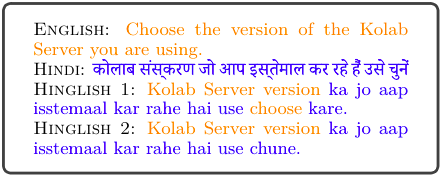}
\caption{Example of the code-mixed sentences generated by the annotator for an \textcolor{orange}{English}-\textcolor{blue}{Hindi} sentence pair.}
\label{fig:example_cms}
\end{figure}

Post annotation, we remove the sentence pairs marked as ``\#'' or are missing an annotation. Note that due to the complexity of generating code-mixed sentences, such as the inability to identify two unique Hinglish sentences, complex sentence structuring, and usage of difficult words in monolingual sentences, annotators do not provide two unique sentences Hinglish sentences for each monolingual sentence pair. We obtain 1,978 sentence pairs with two or more unique Hinglish sentences.
On average, 2.5 code-mixed sentences are created for each Hindi-English sentence pair. Figure \ref{fig:example_cms} shows an example of two code-mixed sentences generated by the annotator for a given sentence pair.

\noindent\textbf{Qualitative evaluation of the human-generated Hinglish text:}
To qualitatively evaluate the generated sentences, we adapt the evaluation strategy described in \cite{srivastava2021challenges}. We randomly sample 100 Hinglish sentences generated by humans along with the source parallel monolingual English and Hindi sentences. We employ two human annotators\footnote{different from the human annotators who generate the Hinglish sentences.} for the qualitative evaluation. We ask the annotators to rate each Hinglish sentence on two metrics:
\begin{itemize}[noitemsep,nolistsep]
    \item \textbf{Degree of code-mixing (DCM)}: The score can vary between 0 to 10. A DCM score of 0 corresponds to the monolingual sentence with no code-mixing, whereas the DCM score of 10 suggests a high degree of code-mixing.
    \item \textbf{Readability (RA)}: RA score can vary between 0 to 10. A completely unreadable sentence due to many spelling mistakes, no sentence structuring, or meaning yields a RA score of 0. A RA score of 10 suggests a highly readable sentence with clear semantics and easy-to-read words.
\end{itemize}

The average DCM scores by the two human annotators are 8.72 and 8.65. The average RA scores are 8.65 and 8.37. The high average scores demonstrate good quality code-mixed sentence generation. Table \ref{tab:example_rating} shows example ratings provided by the two human annotators to the five Hinglish sentences. 

\begin{table*}[!tbh]
\small{
\centering
\resizebox{\hsize}{!}{
\begin{tabular}{|c|c|c|c|c|}
\hline
\multirow{2}{*}{\textbf{Hinglish sentences}} & \multicolumn{2}{c|}{\textbf{Human 1}} & \multicolumn{2}{c|}{\textbf{Human 2}} \\ \cline{2-5} 
& \textbf{DCM}       & \textbf{RA}      & \textbf{DCM}       & \textbf{RA}      \\ \hline
\textcolor{orange}{Media} \textcolor{blue}{ke} \textcolor{orange}{exposure} \textcolor{blue}{ke aadhar par} \textcolor{orange}{Indian} \textcolor{blue}{rajyo ki} \textcolor{orange}{rankings} &    10   &  10     &    8   &    8                          \\ \hline

\textcolor{orange}{You will be more likely to give up before the 30 minutes}, \textcolor{blue}{aap logo se jyada he}.  &    7   &  8     &  9  &  8  \\ \hline

\textcolor{blue}{Shighra hi} maansingh \textcolor{orange}{british} \textcolor{blue}{ke saath saude baazi kar rha tha}.  &   8    &  10     &   8  & 8 \\ \hline

\textcolor{blue}{par} \textcolor{orange}{there's another way}, \textcolor{blue}{aur mai aapko bata kar ja rahi hun}. &   9    &     10   &   9   &  9\\ \hline

“\textcolor{orange}{Aren’t you a tiny bit} \textcolor{blue}{andhvishavaasi}?”  &   9    &  9     &  9   &  9 \\ \hline
\end{tabular}}}
\caption{Example DCM and RA ratings provided by the three human annotators to the human-generated Hinglish sentences. We color code the tokens in the Hinglish sentence based on the language with the scheme: \textcolor{orange}{English} tokens with orange, \textcolor{blue}{Hindi} tokens with blue and language independent tokens with black color.}
\label{tab:example_rating}
\end{table*}

\section{Machine-Generated Hinglish Text}
\label{sec: machine}
In addition to the human-generated code-mixed text, we also generate the Hinglish sentence synthetically by following the Embedded-Matrix theory~\cite{joshi1982processing}. We propose two rule-based Hinglish text generation systems leveraging the parallel monolingual English and Hindi sentences. In both systems, we use Hindi as the matrix language and English as the embedded language. The matrix language imparts structure to the code-mixed text with tokens embedded from embedded language. We also use several linguistic resources in both generation systems. These resources include:

\begin{itemize}
\item \textbf{English-Hindi Dictionary}: We curate 77,805 pairs of English words and the corresponding Hindi meanings from two sources\footnote{\url{http://www.cfilt.iitb.ac.in/~hdict/webinterface\_user/index.php}}\textsuperscript{,}\footnote{\url{https://jankaribook.com/verbs-list-verb-in-hindi/}} to construct an English-Hindi dictionary.
\item
\textbf{Cross-lingual Word Embedding}: We leverage multilingual word vectors for English and Hindi tokens. These word-vectors ($dim=300$) are generated from fastText's multilingual pre-trained model~\cite{bojanowski2017enriching}.   We further map these vectors to a common space using VecMap~\cite{artetxe2018acl}.    

\item
\textbf{GIZA++}: GIZA++ \cite{och03:asc} learns the word alignment between the parallel sentences using an HMM based alignment model in an unsupervised manner. We train GIZA++ on \textit{IIT-B corpus}.

\item
\textbf{Script Transliteration}: We transliterate the code-mixed sentences containing tokens in the Devanagari script to the Roman script~\cite{Ritwik2019}.  

\item
\textbf{YAKE}: We use YAKE~\cite{campos2020yake}, an unsupervised automatic keyword extraction method, to extract the key-phrases from the monolingual English and Hindi sentences.
\end{itemize}

\begin{figure}[!tbh]
\centering
    \includegraphics[width=1.0\linewidth]{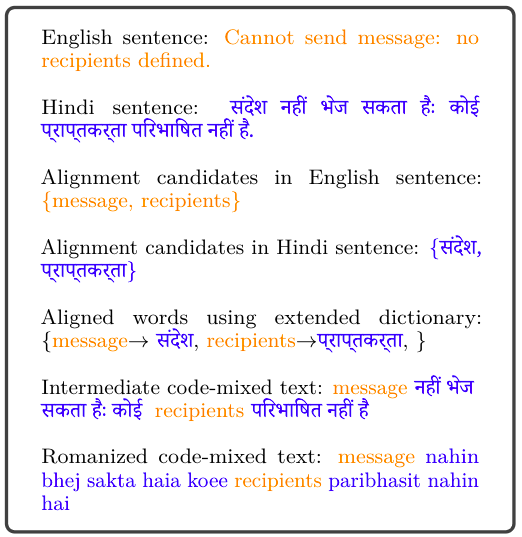}
\caption{An example Hinglish code-mixed sentence generated using WAC method.}
\label{fig:example_wac}
\end{figure}

\begin{figure}[!tbh]
\centering
    \includegraphics[width=1.0\linewidth]{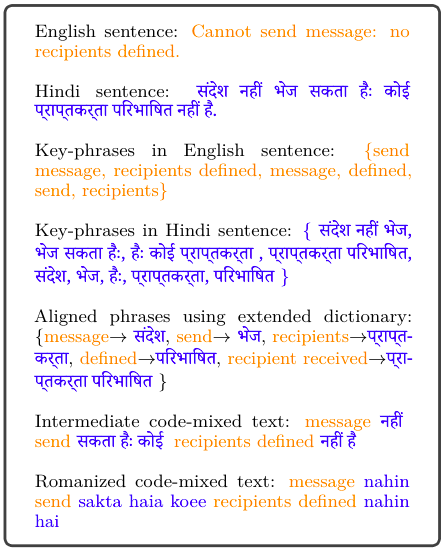}
\caption{An example Hinglish code-mixed sentence generated using PAC method.}
\label{fig:example_pac}
\end{figure}

We further extend the English-Hindi dictionary by incorporating parallel sentences in the \textit{IIT-B corpus}. We leverage VecMap's shared representation to identify the closest word in English and the corresponding Hindi sentence. In addition, we also leverage GIZA++  to align the parallel sentences resulting in aligned English-Hindi tokens. Both of these steps extend the initial dictionary from 77,809 to 1,52,821 words and meaning pairs. We use this extended dictionary in the following two code-mixed text generation systems:
\begin{itemize}
    \item \textbf{Word-aligned code-mixing (WAC)}: Here, we align the noun and adjective tokens between the parallel sentences using the extended dictionary. We replace all the aligned Hindi tokens with the corresponding English noun or adjective token and transliterate the resultant Hindi sentence to the Roman script. Figure \ref{fig:example_wac} demonstrates the example Hinglish text generated from the parallel monolingual English and Hindi sentences using the WAC procedure.
    
    \item \textbf{Phrase-aligned code-mixing (PAC)}: Here, we align the keyphrases of length up to three tokens between the parallel sentences. To identify the keyphrases, we use the YAKE tool. We replace all the aligned Hindi phrases with the corresponding longest matching English phrase and transliterate the resultant Hindi sentence to the Roman script. Figure \ref{fig:example_pac} demonstrates the example Hinglish text generated from the parallel monolingual English and Hindi sentences using the PAC procedure. 
\end{itemize}

\section{A Study on Evaluation of Code-Mixed Text Generation}
\begin{figure}[!tbh]
\centering
    \includegraphics[width=1.0\linewidth]{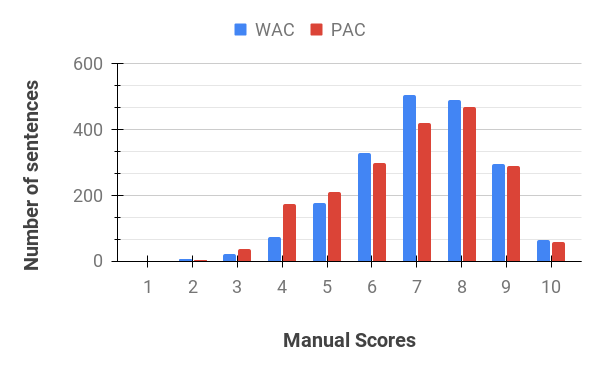}
\caption{Distribution of human evaluation of the generated Hinglish sentences using WAC and PAC.}
\label{fig:ratings}
\end{figure}

\begin{figure}[!tbh]
\centering
    \includegraphics[width=1.0\linewidth]{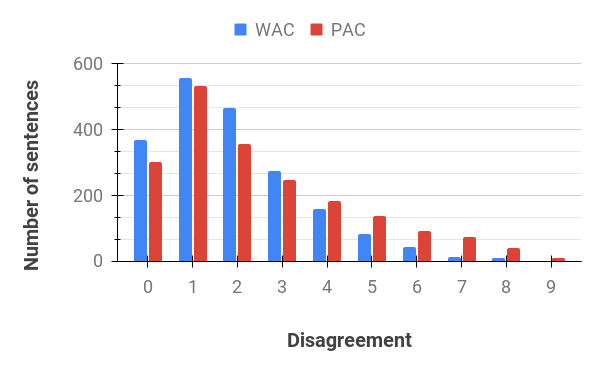}
\caption{Distribution of the disagreement between human evaluation of the generated Hinglish sentences using WAC and PAC.}
\label{fig:disagreement}
\end{figure}

In this section, we evaluate machine-generated code-mixed text. This study demonstrates severe limitations of five widely popular NLP metrics in evaluating code-mixed text generation performance. We leverage the following metrics: (i) Bilingual Evaluation Understudy Score (\textbf{BLEU},~\citet{papineni2002bleu}), (ii)\textbf{NIST}~\cite{doddington2002automatic}, (iii) BERTScore (\textbf{BS},~\citet{zhang2019bertscore}), (iv) Word Error Rate (\textbf{WER},~\citet{levenshtein1966binary}), and (v) Translator Error Rate (\textbf{TER},~\citet{snover2006study}). Higher BLEU, NIST, or BS values and lower WER or TER values represent better generation performance. We conduct two experiments to evaluate the machine-generated Hinglish text:

\begin{itemize}
    \item \textbf{Human evaluation}: First, we perform coarse-grained qualitative evaluation by randomly sampling 100 English-Hindi sentence pairs and corresponding WAC and PAC generated sentences. We employ two human evaluators\footnote{The evaluators are different from the annotators employed in Section~\ref{sec:dataset}).} who are proficient in English and Hindi languages to evaluate the quality of the generated sentences. We ask evaluators to provide one of the two labels --- \textit{Correct} and \textit{Incorrect} --- to each of the generated sentences. A sentence is marked \textit{Correct} if it is following the semantics of the parallel sentences and has high readability. Table \ref{agreement} shows the annotator's agreement on the randomly sampled set of sentences. The coarse-grained qualitative evaluation shows the correct generation in at least 50\% of the cases. Table \ref{tab: performance} shows the evaluation of the randomly sampled 100 sentences on five metrics. The results show the inefficacy of automatic evaluation metrics to capture the linguistic diversity of the generated code-mixed text due to spelling variations, omitted words, limited reference sentences, and informal writing style~\cite{srivastava2020phinc}.
    We further conduct a fine-grained qualitative human evaluation to measure the similarity between the machine-generated Hinglish sentences and the monolingual pair, the readability, and the grammatical correctness. We employ a new set of eight human evaluators to provide a rating between 1 (low quality) to 10 (high quality) to the PAC and WAC generated Hinglish sentences based on the following three parameters:
    \begin{itemize}
     \item The similarity between the generated Hinglish and the monolingual source sentences.
     \item The readability of the generated sentence.
     \item The grammatical correctness of the generated sentence.
    \end{itemize}
    
    Each generated sentence is rated by two human evaluators. All the evaluators have expert-level proficiency in the English and Hindi languages.  Table \ref{fig:instance} shows ratings provided to a representative machine-generated Hinglish sentence.  Figure \ref{fig:ratings} shows the distribution of the fine-grained human evaluation scores. For both procedures, the majority (WAC: 82.3\% and PAC: 75.2\%) of the sentences score in the range 6--9. None of the sentences received a rating score of 1. The results further corroborate our claim that automatic evaluation metrics undermine code-mixed text generation performance. Figure \ref{fig:disagreement} shows the distribution of the disagreement in the human evaluation of the generated sentences. We calculate the disagreement as to the absolute difference between the human evaluation scores. PAC-generated sentences are more prone to high disagreement (>=5) in the human evaluation than WAC. This could be attributed to the fact that PAC-generated sentences are relatively less constrained, which leaves the evaluation to the expertise and interpretation of the Hinglish language by the annotators.  
\end{itemize}

\begin{table}[!tbh]
\centering
\small{
\resizebox{\hsize}{!}{
\begin{tabular}{|c|c|c||c|c|}
\hline
\multirow{2}{*}{}      & \multicolumn{2}{c||}{\textbf{WAC}}     & \multicolumn{2}{c|}{\textbf{PAC}}     \\ \cline{2-5} 
 & \textbf{Agree} & \textbf{Disagree}   & \textbf{Agree} & \textbf{Disagree}   \\ \hline
\textit{\textbf{Correct}}   & 56   & \multirow{2}{*}{22} & 55   & \multirow{2}{*}{40} \\ \cline{1-2} \cline{4-4}
\textit{\textbf{Incorrect}} & 22   &      & 5    &      \\ \hline
\end{tabular}}}
\caption{Annotator agreement on the randomly sampled sentences for WAC and PAC.}
\label{agreement}
\end{table}

\begin{table}[!tbh]
\centering
\small{
\resizebox{\hsize}{!}{
\begin{tabular}{|c|c|c|c|c|c|}
\hline
   & \textbf{BLEU} & \textbf{WER} & \textbf{TER} & \textbf{NIST} & \textbf{BS} \\ \hline
\textbf{WAC} & 0.1229   & 0.8240  & 0.7830  & 2.2045 & 0.857  \\ \hline
\textbf{PAC} & 0.1202   & 0.8228  & 0.7981  & 2.0497 & 0.857  \\ \hline
\end{tabular}
}}
\caption{Automatic performance evaluation of the WAC and PAC procedures.}
\label{tab: performance}
\end{table}

\begin{table*}[!tbh]
\centering
    \includegraphics[width=1.0\linewidth]{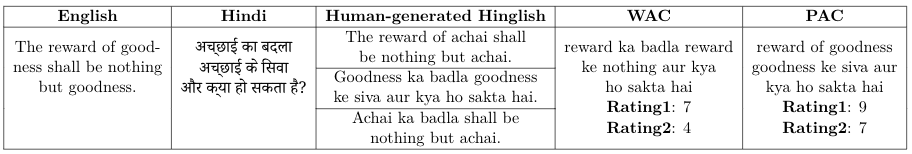}
\caption{Example human-generated and machine-generated Hinglish sentences from the dataset along with the source English and Hindi sentences. Two different human annotators rate the synthetic Hinglish sentences on the scale 1-10 (low-high quality}
\label{fig:instance}
\end{table*}

\begin{itemize}[noitemsep,nolistsep]
    \item \textbf{Metric-based evaluation}: Next, we analyze the performance of five evaluation metrics on the machine-generated code-mixed text. Table \ref{tab:HS} shows the average scores for each of the metrics against the corresponding human-provided rating. As evident from the results, the metrics perform poorly on the code-mixed data. We further analyze the correlation\footnote{We experiment with Pearson Correlation Coefficient. The value ranges from -1 to 1.} of the metric scores with the human-provided ratings for both the text generation procedures. For this task, we divide the human ratings into three buckets:
    \begin{itemize}[noitemsep,nolistsep]
        \item Bucket 1: Human rating between 2--10.
        \item Bucket 2: Human rating between 2--5.
        \item Bucket 3: Human rating between 6--10.
    \end{itemize}
    
    Table \ref{tab:corr} shows the results of the correlation between various metric scores and the human rating to the machine-generated code-mixed sentences using WAC and PAC procedures. Human rating for generated sentences in Bucket 3 is relatively highly correlated with the metric scores compared to Bucket 2. This behavior could be attributed to the fact that low-quality sentences are difficult to rate for the human annotators due to various reasons such as poor sentence structuring and many spelling mistakes.
\end{itemize}

\begin{table*}[!tbh]
\begin{tabular}{|c|ccccc|ccccc|}
\hline
\multirow{2}{*}{\begin{tabular}[c]{@{}c@{}}\textbf{Human} \\ \textbf{Score}\end{tabular}} & \multicolumn{5}{c|}{\textbf{WAC}}     & \multicolumn{5}{c|}{\textbf{PAC}}     \\ \cline{2-11} 
                                                                        & \textbf{BLEU} & \textbf{WER} & \textbf{TER} & \textbf{NIST} & \textbf{BS} & \textbf{BLEU} & \textbf{WER} & \textbf{TER} & \textbf{NIST} & \textbf{BS} \\ \hline

\textbf{2}        & 0.144    & 0.741    &   0.667 &  0.092 & 0.851  
& 0.126    & 0.672   &  0.698 &  0.176 & 0.8603 \\ 

\textbf{3}        & 0.138    & 0.735    &  0.708  & 0.070 & 0.852   
& 0.146    & 0.765    &  0.696  & 0.086 & 0.851 \\ 

\textbf{4}        & 0.133    & 0.695    &  0.666  & 0.103 & 0.849  
& 0.143    & 0.744    &  0.703  & 0.100 & 0.8464  \\ 

\textbf{5}        & 0.135    & 0.711    &  0.681  & 0.110 & 0.853  
& 0.153    & 0.726   &  0.680  & 0.114 & 0.8515  \\ 

\textbf{6}        & 0.141    & 0.697    &   0.670 & 0.102 & 0.852  
& 0.164    & 0.689   &   0.646 & 0.124  & 0.8558  \\ 

\textbf{7}        & 0.161    & 0.663    &   0.630 & 0.111 & 0.856  
& 0.176    & 0.661   &  0.618  & 0.121 & 0.8581   \\ 

\textbf{8}        & 0.177    & 0.621    &   0.589 & 0.127 & 0.859  
& 0.177    & 0.639   &  0.605  & 0.128 & 0.8598  \\ 

\textbf{9}        & 0.212    & 0.571    &   0.538 & 0.150 & 0.865  
& 0.184    & 0.614    &  0.590  & 0.129 & 0.8638 \\ 

\textbf{10}       & 0.291    & 0.509    &   0.493 & 0.157 & 0.878   & 0.242
& 0.551    &   0.543 & 0.146 & 0.8731 \\ \hline

\end{tabular}
\caption{Comparison of various metric scores with the human score for WAC and PAC.}
\label{tab:HS}
\end{table*}

\begin{table}[!tbh]
\resizebox{\hsize}{!}{
\begin{tabular}{|c|cc|cc|cc|}
\hline
\multirow{3}{*}{} 
   & \multicolumn{2}{c|}{\textbf{Bucket 1}}  & \multicolumn{2}{c|}{\textbf{Bucket 2}} & \multicolumn{2}{c|}{\textbf{Bucket 3}}  \\ \cline{2-7} 
   & \textbf{WAC} & \textbf{PAC}  & \textbf{WAC}    & \textbf{PAC}  & \textbf{WAC} & \textbf{PAC}\\ \hline
   
\textbf{BLEU}     & 0.810   & 0.910    & -0.861     &  0.878     & 0.941   & 0.844      \\ 

\textbf{WER} & -0.936  & -0.822    & -0.785     & 0.457     & -0.993  & -0.973    \\ 

\textbf{TER} & -0.891  & -0.963    & 0.000      &  -0.610    & -0.998  & -0.970    \\ 

\textbf{NIST}     & 0.913    & 0.127    & 0.642      & -0.559    & 0.986   & 0.846     \\ 

\textbf{BS} &  0.844  &  0.710   &  0.227   &   -0.689    &   0.953  &  0.937   \\ \hline

\end{tabular}}
\caption{Comparison of correlation between evaluation metrics and human scores for WAC and PAC.}
\label{tab:corr}
\end{table}

\section{Limitations and Opportunities}
In this section, we present a discussion on the various inherent limitations associated with the proposed \textit{HinGE} dataset. We also discuss the various opportunities for the computational linguistic community to build efficient systems and metrics for code-mixed languages. Some of the major limitations with the dataset are:

\begin{itemize}
    \item Due to the high time and cost associated with the human annotations, the number of samples in the dataset is limited. Because of a similar reason, the other code-mixing datasets \cite{srivastava2021challenges} suffer from the scarcity of large-scale human annotations.
    \item The IIT-B parallel corpus does not contain sentences mined from the social media platforms. This potentially reduces the noise in the generated sentences compared to the dataset previously compiled from the social media platforms for various tasks.
    \item The annotators generating the code-mixed sentences were constrained only to include words from the parallel source sentences. This could potentially limit the observations compared to the real-world datasets collected from social media platforms that are more linguistically diverse due to the presence of a large number of multilingual speakers.
    \item Due to the high annotation cost and time, the WAC and PAC generated sentences are only rated on a single scale encompassing multiple dimensions such as grammatical correctness, readability, etc.    
\end{itemize}

Even with the presence of the above limitations, the \textit{HinGE} dataset could be effectively used for various purposes such as:

\begin{itemize}
    \item The dataset could be effectively used in developing code-mixing text generation systems. Currently, the dataset supports only one code-mixed language, i.e., Hinglish, but it could be extended using various techniques such as weak supervision and active learning.
    \item The machine-generated sentences and the corresponding human ratings will be useful in designing metrics and systems for the effective evaluation of various code-mixed NLG tasks. It could also be used to investigate the factors influencing the quality of the code-mixed text.
    \item The dataset could also be used in investigating the reasoning behind the disagreement in the human scores to the machine-generated sentences.
    \item The code-mixed text (human or machine-generated) could be useful in the multitude of other code-mixing tasks such as language identification and POS tagging.
    \item With the recent thrust in code-mixed machine translation, the \textit{HinGE} dataset would be extremely useful in designing and evaluating the machine-translation systems. The multiple code-mixed sentences corresponding to a given pair of parallel monolingual sentences would help to build robust translation systems.
\end{itemize}

\section{Conclusion}
In this paper, we present a high-quality dataset (\textit{HinGE}) for the text-generation and evaluation task in the code-mixed Hinglish language. The code-mixed sentences in the \textit{HinGE} dataset are generated by humans and rule-based algorithms. We demonstrate the poor evaluation capabilities of five widely popular metrics on the code-mixed data.
Along with the human-generated sentences, the machine-generated sentences (as described in Section \ref{sec: machine}) and the human ratings of these code-mixed sentences could facilitate building the highly scalable and robust evaluation metrics and strategies for the code-mixed text. The multiple human-generated sentences corresponding to a pair of parallel monolingual sentences will pave the way in designing natural language generation systems robust to adversaries and linguistic diversities such as spelling variation and matrix language. 

\bibliographystyle{acl_natbib}
\bibliography{custom}
\flushend

\end{document}